\documentclass{article} 
\usepackage{iclr2026_conference,times}

\usepackage{amsmath,amsfonts,bm}

\def\eqref#1{equation~\ref{#1}}

\def\1{\bm{1}}

\DeclareMathAlphabet{\mathsfit}{\encodingdefault}{\sfdefault}{m}{sl}
\SetMathAlphabet{\mathsfit}{bold}{\encodingdefault}{\sfdefault}{bx}{n}

\usepackage{amssymb}
\usepackage{hyperref}
\usepackage{url}
\usepackage{booktabs}
\usepackage{tabularx}
\usepackage{float}

\usepackage{tikz}
\usetikzlibrary{
  positioning,
  fit,
  backgrounds,
  calc,
  shadows.blur
}
\usepackage{framed}
\usetikzlibrary{arrows.meta}

\usepackage{xcolor}

\usepackage[utf8]{inputenc}
\usepackage[T1]{fontenc}

\definecolor{inputblue}{HTML}{DBEAFE}
\definecolor{inputborder}{HTML}{3B82F6}
\definecolor{modelpurple}{HTML}{EDE9FE}
\definecolor{modelborder}{HTML}{7C3AED}
\definecolor{toolgreen}{HTML}{D1FAE5}
\definecolor{toolborder}{HTML}{059669}
\definecolor{submitamber}{HTML}{FEF3C7}
\definecolor{submitborder}{HTML}{D97706}
\definecolor{kernelgray}{HTML}{F3F4F6}
\definecolor{kernelborder}{HTML}{4B5563}
\definecolor{passgreen}{HTML}{059669}
\definecolor{failred}{HTML}{DC2626}
\definecolor{loopcolor}{HTML}{7C3AED}

\definecolor{keywordcolor}{rgb}{0.7, 0.1, 0.1}   
\definecolor{tacticcolor}{rgb}{0.0, 0.1, 0.6}    
\definecolor{commentcolor}{rgb}{0.4, 0.4, 0.4}   
\definecolor{symbolcolor}{rgb}{0.0, 0.1, 0.6}    
\definecolor{sortcolor}{rgb}{0.1, 0.5, 0.1}      
\definecolor{attributecolor}{rgb}{0.7, 0.1, 0.1} 
\definecolor{errorcolor}{rgb}{0.7, 0.1, 0.1}   
\usepackage{listings}

\setcitestyle{round,aysep={,},yysep={,}}
\AtBeginDocument{\raggedbottom\global\vbadness=100000}

\title{FormalProofBench: Can Models Write Graduate Level Math Proofs That Are Formally Verified?}

\author{
\parbox{\textwidth}{\centering
Nikil Ravi$^1$\hspace{2.5em}
Kexing Ying$^2$\hspace{2.5em}
Vasilii Nesterov$^3$\hspace{2.5em}
Rayan Krishnan$^1$ \\[0.6em]
Elif Uskuplu$^4$\hspace{2.5em}
Bingyu Xia$^5$\hspace{2.5em}
Janitha Aswedige$^5$\hspace{2.5em}
Langston Nashold$^1$
}\\[2em]
{\small\centering
\parbox{\textwidth}{\centering
$^1$Vals AI \quad
$^2$\'Ecole Polytechnique F\'ed\'erale de Lausanne (EPFL) \\[0.3em]
$^3$Moscow Institute of Physics and Technology \quad
$^4$Indiana University, Bloomington \\[0.3em]
$^5$Independent Researcher
}}
}

\iclrfinalcopy 
\begin{document}

\maketitle

\begin{abstract}

We present \textbf{FormalProofBench}, a private\footnote{The benchmark is kept private to prevent data contamination. For transparency, we provide sample problems in Appendix~\ref{app:samples}; in addition, the evaluation prompt and all harness code are available at \href{https://github.com/vals-ai/proof-bench}{this https url}.} benchmark designed to evaluate whether AI models can produce formally verified mathematical proofs at the graduate level. Each task pairs a natural-language problem with a Lean~4 formal statement, and a model must output a Lean proof accepted by the Lean 4 checker. FormalProofBench targets advanced undergraduate and graduate mathematics, with problems drawn from qualifying exams and standard textbooks across topics including analysis, algebra, probability, and logic. We evaluate a range of frontier models with an agentic harness, and find that the best-performing foundation model achieves 33.5\% accuracy, with performance dropping rapidly after that. In addition to the accuracy numbers, we also provide empirical analysis of tool-use, failure modes, cost and latency, thereby providing a thorough evaluation of the formal-theorem proving abilities of frontier models.
\end{abstract}

\section{Introduction}

Large language models have made striking progress on mathematical reasoning.
Recent systems achieve high scores on competition-style benchmarks such as
MATH~\citep{hendrycksmath2021}, GSM8K~\citep{cobbe2021trainingverifierssolvemath}, the International Mathematical Olympiad, and more challenging math benchmarks like FrontierMath~\citep{epoch2024thebenchmark}, 
driven by advances in pre-training on mathematical corpora,
post-training with reinforcement learning from human and automated feedback,
and inference-time reasoning.

However, natural-language solutions can hide subtle logical errors that are difficult to audit by both human and LLM-based graders. A model may produce a plausible-sounding argument that omits a critical case or invokes a non-existent lemma, errors that even expert reviewers can overlook. This gap between \emph{apparent} and \emph{verified}  mathematical competence motivates the use of formal verification as an evaluation methodology: a proof either type-checks or it does not.

Independently, the mathematics community has increasingly adopted interactive theorem provers. Lean 4~\citep{moura2021lean4} and its mathematical library Mathlib~\citep{mathlibreference} have experienced rapid growth, and formalization efforts such as the Liquid Tensor Experiment~\citep{commelin2023} and the ongoing formalization of Fermat's Last Theorem~\citep{FLT_Lean} have demonstrated that modern proof assistants can scale to deep, research-level mathematics.  Concurrently, several recent works have explored the intersection of AI and formal mathematics, including Google DeepMind's AlphaProof~\citep{1062014} and the Aristotle system for auto-formalization~\citep{achim2025aristotleimolevelautomatedtheorem}. These developments underscore both the opportunity and the need for rigorous benchmarks that measure whether language models can produce machine-checkable proofs for non-trivial mathematics.

We present \textbf{FormalProofBench}, a private benchmark designed to evaluate formal theorem-proving capability at the advanced undergraduate and graduate level. Each task pairs a natural-language problem statement with a corresponding formal statement in Lean~4; models must produce a proof term accepted by the Lean kernel. Problems are drawn from qualifying exams and standard textbooks spanning measure theory, real and functional analysis, algebra and commutative algebra, algebraic geometry, number theory, probability and stochastic analysis, and logic. All formalizations were written by experienced Lean users, including PhD students and domain experts, and underwent a secondary review process to guard against misformalization.

Unlike informal mathematics benchmarks, FormalProofBench provides a \emph{binary} and \emph{fully automated} correctness signal: a proof either compiles or it does not. This eliminates the need for model-graded or human-graded evaluation of solution correctness and ensures that no credit is awarded for plausible but flawed reasoning.

We evaluate a range of frontier models within an agent harness, providing each model with tools for searching Mathlib, executing Lean code, and submitting proofs, with up to 40 turns of interaction per problem.

\paragraph{Key contributions and takeaways.} 
\begin{enumerate}
    \item We present a new challenging, real-world, deterministic-to-evaluate, and uncontaminated benchmark of language model performance.
    \item We provide a detailed evaluation of frontier models on \textbf{graduate level} formal theorem-proving using 200 carefully formalized problems
    \item We provide an analysis of \textbf{cost and latency} that models use in a \textbf{realistic multi\-turn agentic harness with access to a search tool and a code execution tool}, with up to 40 turns to submit a solution
    \item We find that \textbf{Anthropic's Claude Opus 4.5 (Thinking)} achieves the best performance of 33.5\% accuracy, with performance dropping rapidly after that
    \item Tool use patterns suggest that \textbf{frequent iteration via code execution} correlates with stronger performance
\end{enumerate}

\paragraph{Paper outline.}
Section~\ref{sec:related} surveys related work on mathematical benchmarking and formal theorem proving. Section~\ref{sec:benchmark} describes the benchmark construction, including problem selection, formalization, and quality assurance. Section~\ref{sec:harness} details the evaluation harness and tool interface. Section~\ref{sec:results} presents results and analysis. Section~\ref{sec:limitations} discusses limitations and future directions.

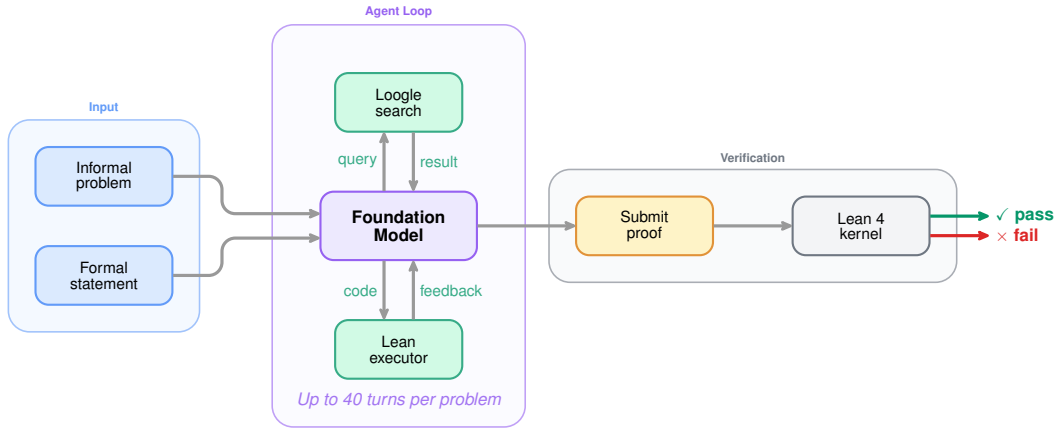
\begin{figure}[t]
\centering
\begin{tikzpicture}[
  scale=0.65, transform shape,
  font=\sffamily\small,
  >={Stealth[length=6pt, width=5pt, round]},
  every node/.style={align=center},
  box/.style={
    rounded corners=5pt, thick,
    minimum height=12mm, minimum width=28mm,
  },
  input/.style={box, draw=inputborder!80, fill=inputblue},
  model/.style={box, draw=modelborder!80, fill=modelpurple, 
    minimum width=32mm, minimum height=14mm,
    font=\sffamily\bfseries},
  tool/.style={box, draw=toolborder!80, fill=toolgreen, minimum width=26mm},
  submit/.style={box, draw=submitborder!80, fill=submitamber},
  kernel/.style={box, draw=kernelborder!80, fill=kernelgray},
  flow/.style={->, very thick, color=black!40},
  biflow/.style={<->, thick, color=toolborder!60},
]
\node[input] (informal) {Informal\\[-1pt]{\footnotesize problem}};
\node[input, below=8mm of informal] (formal) {Formal\\[-1pt]{\footnotesize statement}};
\node[model, right=30mm of $(informal.east)!0.5!(formal.east)$] (fm) {Foundation\\[-1pt]Model};
\node[submit, right=20mm of fm] (submit) {Submit\\[-1pt]{\footnotesize proof}};
\node[kernel, right=16mm of submit] (kernel) {Lean\;4\\[-1pt]{\footnotesize kernel}};
\node[tool, above=12mm of fm] (search) {Loogle\\[-1pt]{\footnotesize search}};
\node[tool, below=12mm of fm] (exec) {Lean\\[-1pt]{\footnotesize executor}};
\begin{scope}[on background layer]
  \node[rounded corners=8pt, fill=inputblue!30, draw=inputborder!48, line width=0.6pt, inner sep=10pt,
        fit=(informal)(formal), 
        label={[font=\sffamily\scriptsize\bfseries, text=inputborder!80]above:Input}] {};
  \node[rounded corners=8pt, fill=modelpurple!20, draw=modelborder!48, line width=0.6pt, inner sep=18pt,
        fit=(search)(exec)(fm), 
        label={[font=\sffamily\scriptsize\bfseries, text=modelborder!80]above:Agent Loop}] {};
  \node[rounded corners=8pt, fill=kernelgray!50, draw=kernelborder!48, line width=0.6pt, inner sep=10pt,
        fit=(submit)(kernel), 
        label={[font=\sffamily\scriptsize\bfseries, text=kernelborder!80]above:Verification}] {};
\end{scope}
\draw[flow, rounded corners=5pt] 
  (informal.east) -- ++(10mm,0) |- ([yshift=2.5mm]fm.west);
\draw[flow, rounded corners=5pt] 
  (formal.east) -- ++(10mm,0) |- ([yshift=-2.5mm]fm.west);
\draw[flow] ([xshift=-3mm]fm.north) -- ([xshift=-3mm]search.south)
  node[midway, left, font=\sffamily\footnotesize, text=toolborder!80] {query};
\draw[flow] ([xshift=3mm]search.south) -- ([xshift=3mm]fm.north)
  node[midway, right, font=\sffamily\footnotesize, text=toolborder!80] {result};
\draw[flow] ([xshift=-3mm]fm.south) -- ([xshift=-3mm]exec.north)
  node[midway, left, font=\sffamily\footnotesize, text=toolborder!80] {code};
\draw[flow] ([xshift=3mm]exec.north) -- ([xshift=3mm]fm.south)
  node[midway, right, font=\sffamily\footnotesize, text=toolborder!80] {feedback};
\draw[flow] (fm.east) -- (submit.west);
\draw[flow] (submit.east) -- (kernel.west);
\draw[->, very thick, color=passgreen] 
  ([yshift=2mm]kernel.east) -- ++(12mm,0) 
  node[right, font=\sffamily\bfseries, text=passgreen] {\checkmark\;pass};
\draw[->, very thick, color=failred] 
  ([yshift=-2mm]kernel.east) -- ++(12mm,0) 
  node[right, font=\sffamily\bfseries, text=failred] {$\times$\;fail};
\node[font=\sffamily\normalsize\itshape, text=loopcolor!70,
      anchor=north, inner sep=2pt]
  at ([yshift=-2mm]exec.south) {Up to 40 turns per problem};
\end{tikzpicture}
\caption{\textbf{Evaluation pipeline for FormalProofBench.} Models are given the informal problem statement, along with the formalized statement (+ necessary headers and context), and have 40 turns (with each turn potentially involving multiple tool calls and reasoning), to submit a proof.}
\label{fig:agent}
\end{figure}

\section{Related Work}\label{sec:related}

\paragraph{Informal mathematics benchmarking}

Natural language mathematics (which we refer to as ``informal'') have been widely used in evaluating and guiding LLM training. Two widely adopted benchmarks are MATH \citep{hendrycksmath2021}, which consists of 12,500 competition-mathematics problems, and AIME, a popular benchmark measuring how well models do on American Invitational Mathematics Competition problems. More recently, benchmarks such as IMProofBench and FrontierMath have begun evaluating models on harder, research-level tasks.

\paragraph{Formal mathematics benchmarks}

Several benchmarks evaluate neural theorem provers at different difficulty levels.
MiniF2F \citep{zheng2022minif2fcrosssystembenchmarkformal} is a cross-system benchmark containing olympiad-level problems formalized in multiple proof assistants.
PutnamBench \citep{tsoukalas2024putnambenchevaluatingneuraltheoremprovers} extends this to collegiate competition mathematics, providing Putnam problems in Lean, Coq, and Isabelle.
ProofNet \citep{azerbayev2023proofnetautoformalizingformallyproving} targets undergraduate-level theorem proving and autoformalization in Lean 3.
FormalMATH \citep{yu2025formalmathbenchmarkingformalmathematical} offers a larger-scale benchmark that also addresses undergraduate mathematics.

Domain-specific benchmarks address particular mathematical subfields.
IneqComp \citep{zhao2025ineqcompbenchmarkinghumanintuitivecompositional} tests compositional reasoning with mathematical inequalities.
CombiBench \citep{liu2025combibenchbenchmarkingllmcapability} focuses on combinatorial mathematics with 100 competition problems formalized in Lean 4.
MATPBench \citep{he2025matpbenchmllmgoodautomated} evaluates the capability of models in multi-modal reasoning in a formal setting.

\paragraph{Formal theorem proving}

On the methods side, autoformalization with LLMs \citep{wu2022autoformalizationlargelanguagemodels} demonstrated that large language models can translate natural-language mathematics into formal specifications.
GoedelProver \citep{lin2025goedelproverfrontiermodelopensource} builds on this by training iteratively on autoformalized proofs, achieving strong results on MiniF2F and PutnamBench.
LeanDojo v2 \citep{hsiang2025leandojov} provides an end-to-end framework for AI-assisted theorem proving in Lean 4, unifying repository extraction, LLM fine-tuning, and proof search.
Lean Workbook \citep{ying2025leanworkbooklargescalelean} provides approximately 57,000 formal-informal problem pairs, including IMO questions, generated via an iterative translation pipeline.
SorryDB \citep{letson2026sorrydbaiproverscomplete} indexes incomplete proofs (``sorries'') from public Lean repositories, serving as a live benchmark based on publicly available unresolved sorries.

FormalProofBench aims to evaluate models' formal theorem-proving abilities on tasks that are at the advanced undergraduate or graduate level, as opposed to high-school problems. In addition, we evaluate models with an agentic scaffold, instead of evaluating their ability to generate the proof in one single forward pass.

\section{Benchmark overview}\label{sec:benchmark}

Each task in FormalProofBench consists of 
(i) a mathematical problem in natural-language, and
(ii) the corresponding formal statement in Lean~4.
Given both, a model must construct a Lean proof accepted by the proof checker.

\subsection{Data Curation}

Problems were selected and formalized by a total of six formalizers who were PhD students or experts experienced in writing Lean 4 code. All of them had significant prior experience writing Lean, either via merged contributions to Mathlib, or prior large-scale formalization effort that they contributed to, and had varied mathematical interests, enabling us to formalize problems from different fields. Our problem-collection process used the following criteria:
\begin{itemize}
    \item Problems that required knowledge beyond olympiad-style / high-school style questions.
    \item Problems that could be formalized using current Mathlib, and where the proof could also be written using current Mathlib.
    \item Problems that would challenge a reasonably capable model as of late 2025, while still being relevant to mathematical work in that domain.
    \item A diversity of mathematical domains.
\end{itemize}

For examples of a task see Listing~\ref{lst:formalization}; 
full model responses for this problem and others are given in 
Appendix~\ref{app:samples}.

\textbf{Problem sources and domains} The problems are drawn from advanced undergraduate and graduate sources, including qualifying exams and standard textbooks. They span a broad range of modern mathematics, including measure theory, real and
functional analysis, algebra and commutative algebra, algebraic geometry, number theory, probability and stochastic analysis, and logic/model theory. Table~\ref{tab:subject-breakdown} presents the distribution of the 200 benchmark problems across these areas.

\begin{table}[ht]
\centering
\begin{tabular}{lr}
\toprule
\textbf{Subject Area} & \textbf{Count} \\
\midrule
Real Analysis & 56 \\
Probability Theory / Stochastic Analysis & 34 \\
Functional Analysis & 26 \\
Number Theory & 23 \\
Logic / Model Theory & 21 \\
Algebra & 18 \\
Algebraic Geometry & 15 \\
Linear Algebra & 7 \\
\midrule
\textbf{Total} & \textbf{200} \\
\bottomrule
\end{tabular}
\caption{Breakdown of FormalProofBench problems by subject area.}
\label{tab:subject-breakdown}
\end{table}

\textbf{Data quality and misformalization checks} While Lean 4 can act as a checker of whether a proof indeed proves a statement, a common problem in formal math benchmarks is misformalization of the \textit{problem statement}, due to misaligned definitions, or incorrect translation to Lean 4. In other words, it is not guaranteed whether the formal statement faithfully expresses the intent of the informal statement. In addition, it is well-known that formal benchmarks often have issues, including misformalized problems and omitted side conditions \citep{ammanamanchi2025faults}.

Anticipating such issues, our data collection process included a secondary review phase of human review and revisions to problem statements. The secondary review phase was performed using standard code review processes on GitHub, and involved a review of the informal mathematics statement, natural language proof, and the formal statement by one or more people other than the original contributor of the problem. 

\textbf{Necessity of a search tool} Some problems in the benchmark have solutions that depend on a theorem that was recently added to \texttt{mathlib}. Although models could in principle reprove any needed lemmas, we include these problems to test a crucial capability for real-world formalization: reliably looking up relevant theorems.

\begin{lstlisting}[
  float,
  frame=single,
  label={lst:formalization},
  caption={A sample formalization from the benchmark, drawn from Dembo's probability theory lecture notes~\cite{dembo2024probability} (Exercise 1.1.5).}
]
import Mathlib

open TopologicalSpace Filter MeasureTheory ProbabilityTheory Function

open scoped NNReal ENNReal MeasureTheory ProbabilityTheory Topology

namespace MeasureTheory

variable {Omega : Type*} [MeasurableSpace Omega]

/-- Prove that a finitely additive non-negative set function mu on
a measurable space (Omega, F) with the "continuity" property:
Bn in F, Bn -> empty, mu(Bn) < inf => mu(Bn) -> 0 must be
countably additive if mu(Omega) < inf. -/
theorem dembo_1_1_15 {mu : Set Omega -> Real}
    (hAdd : forall A B, MeasurableSet A -> MeasurableSet B ->
      Disjoint A B -> mu (A \cup B) = mu A + mu B)
    (hCont : forall B : Nat -> Set Omega,
      (forall n, MeasurableSet (B n)) ->
      Antitone B -> (Inter n, B n) = empty ->
      Tendsto (fun n => mu (B n)) atTop (NHDS 0))
    {A : Nat -> Set Omega} (hA : forall n, MeasurableSet (A n))
    (hDisj : Pairwise (Disjoint on A)) :
    mu (Union n, A n) = tsum n, mu (A n) := by
  sorry

end MeasureTheory
\end{lstlisting}

\subsection{Evaluation settings}

A common problem with evaluations is that different models have different settings, such as reasoning effort, temperature, $\texttt{top\_p}$, inference provider, etc. We standardize our evaluation settings by using the default sampling settings from each model provider, and by using the native model provider rather than a third-party inference provider.

\section{Evaluation harness and tools}\label{sec:harness}

We provide models with the natural language problem statement, and the formalization of this statement, along with the necessary headers and context. We also provide the following tools:
\begin{itemize}
    \item \texttt{lean\_loogle}: A tool that models can use to search for theorems and definitions in Mathlib, via https://loogle.lean-lang.org.
    \item \texttt{lean\_run\_code}: Models can use this tool to execute any Lean code and get feedback, including whether it succeeded or failed, error descriptions, unsolved goals, type mismatches, warnings, \texttt{sorry}s, syntax errors, missing definitions/lemmas, and more.
    \item \texttt{submit\_proof}: Used to indicate final submission of proof, starting with \texttt{by}.
\end{itemize}

In order to allow models to iterate on their solution before submitting, each problem allows up to 40 turns. A solution is considered valid if and only if the model calls \texttt{submit\_proof} with a proof that successfully passes the Lean 4 kernel. We use Lean \textbf{v4.25.2} and Mathlib (at commit \texttt{c98ae54af00eaefe79c51b2b278361ca94e59bfb}). We also reject any attempt containing \texttt{axiom} or \texttt{local\_instance}, since that would circumvent formal verification.

\begin{figure}[H]
\centering
\begin{minipage}{0.85\linewidth}
\scriptsize
\begin{framed}
\begin{verbatim}
SYSTEM_PROMPT = """
You are a Lean 4 theorem proving expert and an expert in graduate level
mathematics. You will be given a graduate level mathematics theorem statement
and its Lean 4 formalization. You must output the Lean 4 proof (your output
must only have the final proof; just the part starting with 'by').
Pay attention to the way the theorem statement is formalized; there may be
specific nuances in the way the statement has been formalized. This will be
important for you to understand the statement and prove it formally.
You will have access to a set of tools to help you prove the theorem. You
may rely on these tools if needed; be smart about how and when you use them.
You may use these tools however you please within the constraints of the
budget and turns, but you must call submit_proof with the final proof (the
part starting with 'by') when done.
Use valid, correct Lean 4 syntax; the tools may be helpful if in doubt.
Key Lean 4 syntax rules:
- Use CamelCase for lemmas (e.g., Nat.dvd_mul, not nat.dvd_mul)
- Chain tactics with ';' not ','
- Rewrites require brackets: rw [h] or rw [step1, step2]
- Use 'constructor' for conjunctions, not 'split'
- Use 'ring' for ring axioms, 'linarith' for linear arithmetic
- Existential witnesses: use 1, 2, 3 not use [1, 2, 3]
- No indentation on tactics
- In your final submission, never use 'sorry', even in comments
  or in the last step.
"""
\end{verbatim}
\end{framed}
\end{minipage}
\caption{System prompt used for all model evaluations.}
\label{fig:system-prompt}
\end{figure}

\begin{figure}[H]
\centering
\begin{minipage}{0.85\linewidth}
\scriptsize
\begin{framed}
\begin{verbatim}
USER_PROMPT_TEMPLATE = """Prove the following theorem in Lean 4.
Natural Language Statement:
{natural}
Formal Statement:
```lean
{header}
{formal}
```
{tool_guidance}
Provide a proof starting with `by`. You must call submit_proof
when you want to submit your final proof."""
\end{verbatim}
\end{framed}
\end{minipage}
\caption{User prompt template. Variables in braces are filled per-problem. Tool guidance  is appended via \texttt{\{tool\_guidance\}}. See Figure~\ref{fig:tool-guidance} for the full tool guidance prompt.}
\label{fig:user-prompt}
\end{figure}

\section{Results}\label{sec:results}

Rather than evaluating several smaller open-source models, we chose to evaluate stronger frontier foundation models, with the goal of benchmarking the most capable models today. Table~\ref{tab:proofbench-results} summarizes the overall results on models from OpenAI, Anthropic, xAI, Zhipu AI, Google, and DeepSeek.

\begin{table}[H]
\centering
\small
\setlength{\tabcolsep}{4pt}
\begin{tabularx}{\linewidth}{@{}rXrrr@{}}
\toprule
\textbf{Rank} & \textbf{Model} & \textbf{Acc. (\%)} & \textbf{Cost / test (\$)} & \textbf{Latency (s) / test}\\
\midrule
1  & Claude Opus 4.5 (Thinking)         & 33.50 & 1.34 & 4036.78\\
2  & Gemini 3 Pro (11/25)               & 18.50 & 1.26 & 9966.43\\
3  & Claude Sonnet 4.5 (Thinking)       & 18.00 & 0.84 & 11186.79\\
4  & GPT 5                              & 17.00 & 0.89 & 5017.28\\
5  & Gemini 3 Flash (12/25)             & 15.00 & 0.33 & 1305.37\\
6  & GPT 5.2                            & 13.50 & 1.10 & 1459.24\\
7  & GPT 5 Nano                         & 12.50 & 0.05 & 3994.69\\
8  & GPT 5.1 Codex Max                  & 12.00 & 0.55 & 2552.82\\
9  & DeepSeek V3.2 (Nonthinking)        & 8.00  & 0.12 & 3715.92\\
10 & GPT 5 Mini                         & 7.00  & 0.21 & 2805.07\\
11 & GLM 4.7                            & 4.50  & 0.15 & 3403.63\\
12 & Grok 4.1 Fast (Reasoning)          & 4.00  & 0.10 & 1455.68\\
13 & DeepSeek V3.2 (Thinking)           & 3.00  & 0.13 & 3808.62\\
\bottomrule
\end{tabularx}
\caption{FormalProofBench overall leaderboard results (accuracy (in \%), cost per test, and latency per test).}
\label{tab:proofbench-results}
\end{table}

\subsection{Tool-use analysis}

\begin{figure}[H]
\centering
\includegraphics[width=\linewidth]{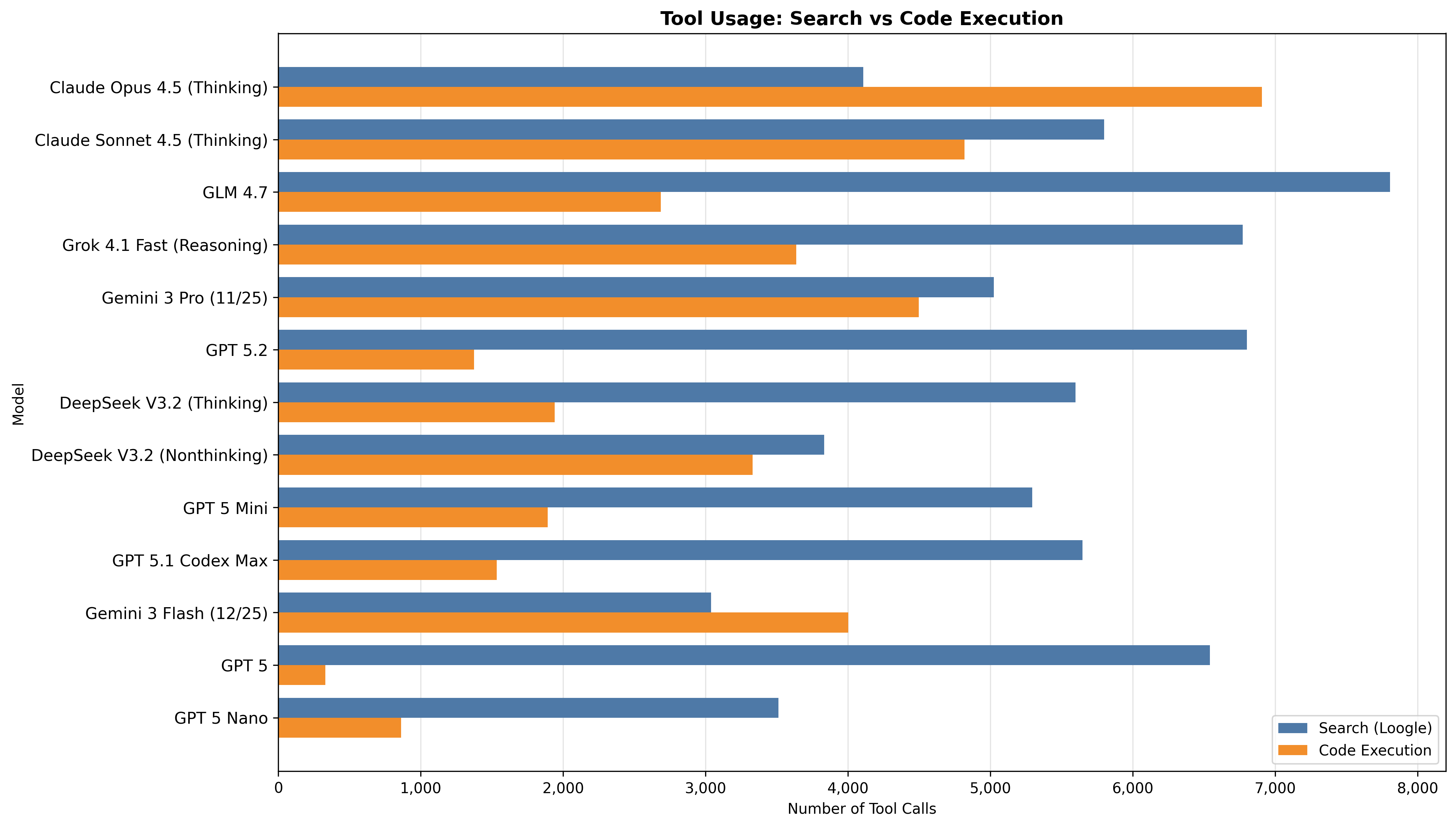}
\caption{Total tool calls (search vs.\ code execution) aggregated across 
all 200 problems per model. Models vary widely in their tool-use strategies.}
\label{fig:tool-use}
\end{figure}

A common failure mode is excessive search: models sometimes spiral into repeated
\texttt{lean\_loogle} tool calls while attempting to find non-existent lemmas.
In contrast, stronger models tend to start executing code early and use search to fill gaps mid-proof.
We observe that higher rates of code execution feedback correlate with stronger
overall success. The qualitative reason for this is that models that tend to execute code get more feedback, and can therefore iterate towards the correct solution, only using the search tool when they actually need it (see Figure~\ref{fig:tool-use}).

\begin{figure}[H]
\centering
\includegraphics[width=\linewidth]{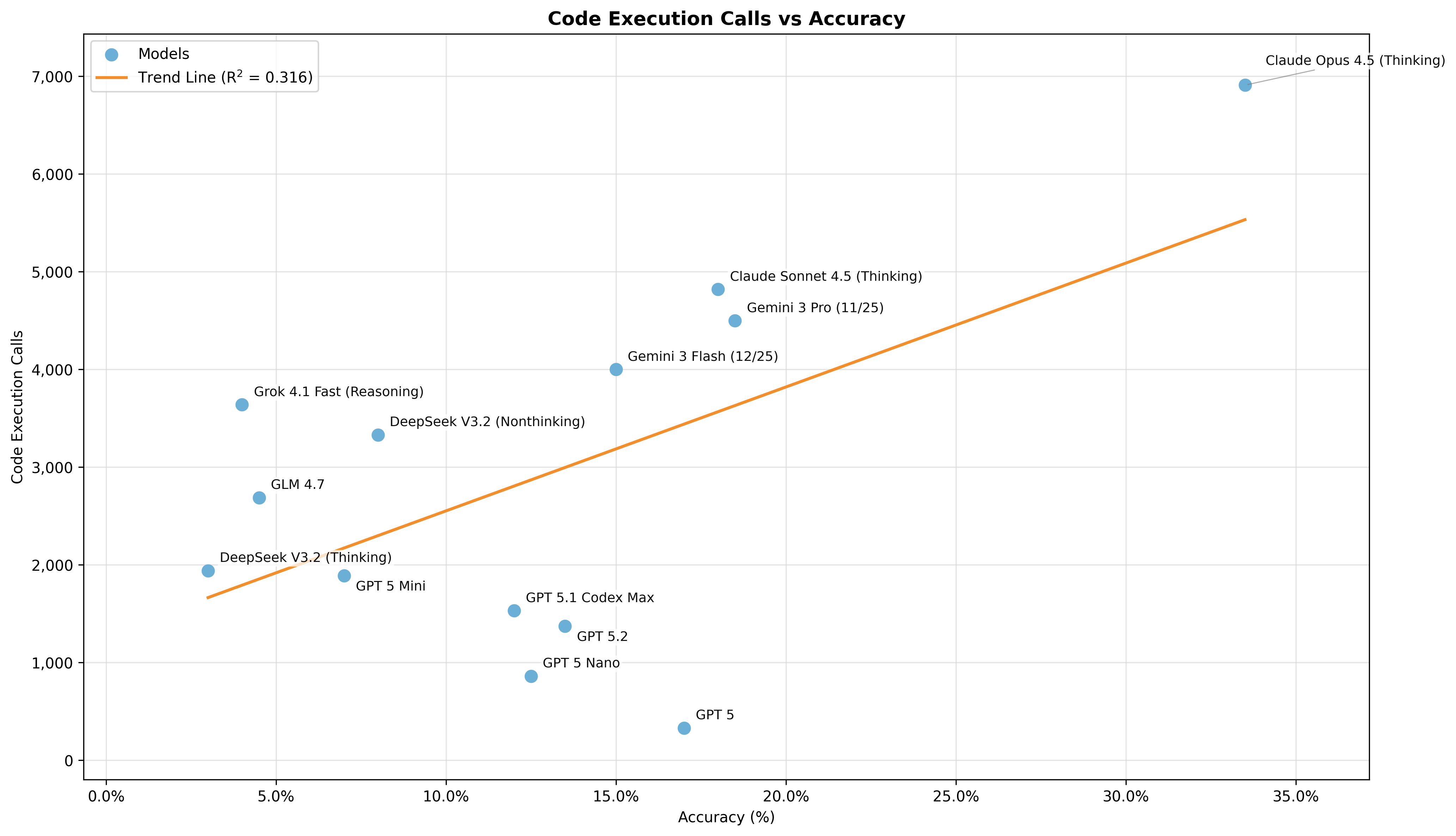}
\caption{We observe a positive but modest correlation between the number of code execution calls and accuracy across models. Notable outliers include GPT~5, which achieves 
relatively high accuracy with very few code execution calls, and Grok~4.1 Fast, which executes substantial code but 
achieves low accuracy.}
\label{fig:other-figure}
\end{figure}

\subsection{Latency and Cost}

In addition to measuring accuracy, it is important to also measure the cost and latency associated with each model's accuracy. Latency is measured as wall-clock time for the entire query execution (including retries), and cost is calculated from token usage and model-specific pricing rules; we also incorporate caching in our cost calculation. The average cost per test ranges from around 10 cents to more than a dollar. The latency in Table~\ref{tab:proofbench-results} indicates that models often take more than an hour (per-test average) when attempting to arrive at their final submission in this benchmark.

\section{Conclusion and future work}\label{sec:limitations}

We present FormalProofBench, a benchmark evaluating whether AI models can produce formally verified mathematical proofs at the graduate level. Our evaluation of frontier models reveals that the best-performing system, Claude Opus 4.5 (Thinking), achieves 33.5\% accuracy, with performance dropping rapidly thereafter. Through detailed analysis of tool use patterns, we find that frequent iteration via code execution correlates strongly with success, while excessive search without execution often indicates a failing trajectory.

\paragraph{A benchmark at the intersection of multiple desiderata.}
FormalProofBench occupies a unique position in the landscape of AI evaluation. Recent benchmarks such as FrontierMath prioritize including challenging problems of real world significance. FormalProofBench does this but also offers several additional advantages that are difficult to achieve simultaneously: (i) deterministic evaluation via the Lean 4 kernel eliminates the need for model-graded or human-graded correctness checks, (ii) contamination resistance, as the specific Lean formalizations we created are unlikely to appear in training data, (iii) practical evaluation costs, with per-problem evaluation completing in minutes to hours rather than days, and (iv) solvability, since we chose problems that can be correctly stated/solved in Lean using current Mathlib. This combination makes FormalProofBench particularly valuable for tracking progress as models continue to improve.

\paragraph{Implications for mathematical practice.}
The gap between informal and formal mathematical competence reveals that formalization remains a significant bottleneck. However, the rapid pace of improvement across recent model releases suggests we may be approaching a threshold where AI systems become practical tools for working mathematicians. Success at formal verification could enable rapid exploration of proof strategies with immediate correctness feedback, democratize access to formal methods beyond the small community of proof assistant experts, and accelerate the formalization of research-level mathematics. As performance on benchmarks like FormalProofBench continues to improve, formal verification may transition from a specialized technique to a standard component of the mathematical research toolkit.

\paragraph{Future directions.}
Several extensions would deepen our understanding of model capabilities:
\begin{enumerate}
    \item \textbf{Exploring alternative tool configurations:} The \texttt{lean\_loogle} tool was significantly less useful to models than the \texttt{lean\_run\_code} tool, and we observed that there were a non-trivial number of trajectories where models unsuccessfully over-used the Loogle tool. Alternative configurations including semantic search, the ability to \texttt{grep} local codebases, and tactic-level tools may reveal whether the current search interface is the bottleneck or whether deeper reasoning limitations persist.
    \item \textbf{Ablating the informal problem statement:} We currently provide both the natural language statement and formal statement to models. Analyzing how performance changes with and without informal proof access would clarify whether models struggle primarily with proof generation or with translating mathematical ideas into Lean syntax.
    \item \textbf{Broader diversity and coverage:} Our benchmark consists of 200 problems. Expanding coverage to additional mathematical domains and increasing problem count would improve statistical power and enable domain-specific analysis.
    \item \textbf{Evaluating research-level formal theorem proving:} While FormalProofBench considers advanced undergraduate and graduate level mathematics, we were constrained by the current contents of \texttt{Mathlib} in what we could plausibly formalize. As more mathematics becomes formalized and available in \texttt{Mathlib}, future iterations of this benchmark could target research-level theorems, providing a pathway to evaluate AI systems on problems at the frontier of mathematical knowledge.
\end{enumerate}

\bibliography{iclr2026_conference}
\bibliographystyle{iclr2026_conference}

\appendix
\section{Appendix}

\subsection{Sample Problems and Model Responses}\label{app:samples}

We present sample problems from the benchmark along with responses from two models, illustrating the difference between a failed and a successful proof attempt. 

\paragraph{Problem 1.} Show that an $A$-scheme is separated (over $A$) if and only if it is separated over $\mathbb{Z}$. This problem is drawn from Vakil's The Rising Sea~\citep{vakil2025rising} (Chapter 11.2, Exercise E).
{\small
\begin{lstlisting}[
  frame=single,
  label={lst:vakil-formalization},
  caption={Lean 4 formalization of the separatedness problem from Vakil Ch.~11.2.}
]
import Mathlib

open AlgebraicGeometry

variable {A : Type*} {X : Scheme} [CommRing A]
    (struct_X : X --> Spec (CommRingCat.of A))

theorem algebraicGeometry_vakil_ch_11_2_E :
    IsSeparated struct_X <-> IsSeparated (instOverTerminalScheme X).hom := by
  sorry
\end{lstlisting}}

\paragraph{GPT 5.2 (Correct).} The following proof successfully compiles. It explicitly constructs the \texttt{IsSeparated} instances via \texttt{IsSeparated.of\_comp} and manages the composition of morphisms to the terminal object.
{\small
\begin{lstlisting}[
  frame=single,
  label={lst:gpt52-response},
  caption={GPT 5.2 response (correct): compiles successfully.}
]
by
  simpa using
    (show IsSeparated struct_X <-> IsSeparated
        (CategoryTheory.Limits.terminal.from X) from by
      constructor
      * intro h
        have hterm :
            IsSeparated
              (CategoryTheory.Limits.terminal.from
                (Spec (CommRingCat.of A))) := by
          infer_instance
        have hcomp :
            IsSeparated
              (CategoryTheory.CategoryStruct.comp struct_X
                (CategoryTheory.Limits.terminal.from
                  (Spec (CommRingCat.of A)))) :=
            (CategoryTheory.MorphismProperty.IsStableUnderComposition.comp_mem
            (P := @AlgebraicGeometry.IsSeparated)
            struct_X
            (CategoryTheory.Limits.terminal.from
              (Spec (CommRingCat.of A))) h hterm)
        simpa [CategoryTheory.Limits.terminal.comp_from struct_X]
          using hcomp
      * intro h
        have hcomp :
            IsSeparated
              (CategoryTheory.CategoryStruct.comp struct_X
                (CategoryTheory.Limits.terminal.from
                  (Spec (CommRingCat.of A)))) := by
          simpa [CategoryTheory.Limits.terminal.comp_from struct_X]
            using h
        haveI :
            IsSeparated
              (CategoryTheory.CategoryStruct.comp struct_X
                (CategoryTheory.Limits.terminal.from
                  (Spec (CommRingCat.of A)))) :=
          hcomp
        exact
          AlgebraicGeometry.IsSeparated.of_comp struct_X
            (CategoryTheory.Limits.terminal.from
              (Spec (CommRingCat.of A))))
\end{lstlisting}}

\paragraph{Claude Opus 4.5 Thinking (Incorrect).} The following proof attempt has the right high-level structure but fails due to instance resolution issues with \texttt{IsSeparated}.
{\small
\begin{lstlisting}[
  frame=single,
  label={lst:opus-response},
  caption={Claude Opus 4.5 Thinking response (incorrect): fails due to instance resolution issues.}
]
by
  have key : struct_X >> Limits.terminal.from
      (Spec (CommRingCat.of A)) =
      (instOverTerminalScheme X).hom := by
    apply Limits.terminal.hom_ext
  constructor
  * intro h
    rw [<- key]
    exact inferInstance
  * intro h
    have h' : IsSeparated (struct_X >>
        Limits.terminal.from (Spec (CommRingCat.of A))) := by
      rw [key]; exact h
    exact IsSeparated.of_comp struct_X
      (Limits.terminal.from (Spec (CommRingCat.of A)))
\end{lstlisting}}

This example illustrates a common failure pattern: Claude Opus 4.5 (Thinking) identifies the correct proof strategy, showing that the two morphisms are related via composition with the terminal morphism, but fails at the level of Lean's type class resolution. GPT 5.2 succeeds by explicitly constructing the required instances and carefully managing the \texttt{IsSeparated} typeclass through composition and simplification lemmas.

\paragraph{Problem 2.} Prove that a finitely additive non-negative set function $\mu$ on a measurable space $(\Omega, \mathcal{F})$ with the ``continuity'' property $B_n \in \mathcal{F}$, $B_n \downarrow \varnothing$, $\mu(B_n) < \infty \implies \mu(B_n) \to 0$ must be countably additive if $\mu(\Omega) < \infty$. This problem is taken from Amir Dembo's probability theory lecture notes~\citep{dembo2024probability} (Exercise 1.1.5) and given in Listing~\ref{lst:formalization}.

\paragraph{Claude Opus 4.5 Thinking (Correct).} The following proof successfully compiles. It establishes countable additivity by decomposing $\bigcup_i A_i$ into a finite union and a tail $B_n = \bigcup_{k \geq n} A_k$, showing $B_n \downarrow \varnothing$ and $\mu(B_n) \to 0$, then concluding via \texttt{NNReal.hasSum\_iff\_tendsto\_nat}.

{\small
\begin{lstlisting}[
  frame=single,
  label={lst:opus-dembo},
  caption={Claude Opus 4.5 Thinking response (correct): compiles successfully.}
]
by
  have hEmpty : mu empty = 0 := by
    have h := hAdd empty empty MeasurableSet.empty
      MeasurableSet.empty disjoint_bot_left
    simp only [Set.union_empty] at h
    have : mu empty + mu empty = mu empty := h.symm
    have h3 : mu empty + mu empty - mu empty =
        mu empty - mu empty := by rw [this]
    simp at h3; exact h3
  set B := fun n => iUnion k >= n, A k with hBdef
  have hB_meas : forall n, MeasurableSet (B n) := fun n =>
    MeasurableSet.biUnion (Set.to_countable _)
      (fun k _ => hA k)
  have hB_anti : Antitone B := by
    intro m n hmn
    apply Set.biUnion_subset_biUnion_left
    intro k hk; exact le_trans hmn hk
  have hB_inter : iInter n, B n = empty := by
    rw [Set.eq_empty_iff_forall_not_mem]
    intro x hx
    simp only [Set.mem_iInter] at hx
    have h0 := hx 0; rw [hBdef] at h0
    simp only [Set.mem_iUnion, ge_iff_le] at h0
    obtain <k, -, hxk> := h0
    have hk1 := hx (k + 1); rw [hBdef] at hk1
    simp only [Set.mem_iUnion, ge_iff_le] at hk1
    obtain <j, hjk, hxj> := hk1
    have hkj : k /= j := by omega
    exact Set.disjoint_iff.mp (hDisj hkj) <hxk, hxj>
  have hBtends : Tendsto (fun n => mu (B n)) atTop (NHDS 0) :=
    hCont B hB_meas hB_anti hB_inter
  have hUnion : forall n, (iUnion i, A i) =
      (iUnion k < n, A k) \cup B n := by
    intro n; ext x
    simp only [Set.mem_union, Set.mem_iUnion, exists_prop, hBdef]
    constructor
    * rintro <k, hxk>
      by_cases hkn : k < n
      * left; exact <k, hkn, hxk>
      * right; exact <k, not_lt.mp hkn, hxk>
    * rintro (<k, -, hxk> | <k, -, hxk>) <;> exact <k, hxk>
  have hDisjParts : forall n,
      Disjoint (iUnion k < n, A k) (B n) := by
    intro n; rw [Set.disjoint_iff]
    intro x <hx1, hx2>
    simp only [Set.mem_iUnion, exists_prop, hBdef] at hx1 hx2
    obtain <k1, hk1n, hxk1> := hx1
    obtain <k2, hk2n, hxk2> := hx2
    have hne : k1 /= k2 := by omega
    exact Set.disjoint_iff.mp (hDisj hne) <hxk1, hxk2>
  have hFinMeas : forall n,
      MeasurableSet (iUnion k < n, A k) := fun n =>
    MeasurableSet.biUnion (Set.to_countable _)
      (fun k _ => hA k)
  have hFinAdd : forall n,
      mu (iUnion k < n, A k) =
      sum k in Finset.range n, mu (A k) := by
    intro n
    induction n with
    | zero => simp [hEmpty]
    | succ n ih =>
      rw [Finset.sum_range_succ, <- ih]
      have hunion : (iUnion k < n + 1, A k) =
          (iUnion k < n, A k) \cup A n := by
        ext x; simp only [Set.mem_iUnion, Set.mem_union,
          exists_prop]
        constructor
        * rintro <k, hkn, hxk>
          by_cases hkn' : k < n
          * left; exact <k, hkn', hxk>
          * have heq : k = n := by omega
            right; rw [heq] at hxk; exact hxk
        * rintro (<k, hkn, hxk> | hxn)
          * exact <k, Nat.lt_succ_of_lt hkn, hxk>
          * exact <n, Nat.lt_succ_self n, hxn>
      have hdisj : Disjoint (iUnion k < n, A k) (A n) := by
        rw [Set.disjoint_iff]; intro x <hx1, hx2>
        simp only [Set.mem_iUnion, exists_prop] at hx1
        obtain <k, hkn, hxk> := hx1
        have hne : k /= n := by omega
        exact Set.disjoint_iff.mp (hDisj hne) <hxk, hx2>
      rw [hunion]; exact hAdd _ _ (hFinMeas n) (hA n) hdisj
  have hKey : forall n, mu (iUnion i, A i) =
      (sum k in Finset.range n, mu (A k)) + mu (B n) := by
    intro n
    calc mu (iUnion i, A i)
        = mu ((iUnion k < n, A k) \cup B n) := by
          rw [hUnion n]
      _ = mu (iUnion k < n, A k) + mu (B n) :=
          hAdd _ _ (hFinMeas n) (hB_meas n) (hDisjParts n)
      _ = (sum k in Finset.range n, mu (A k)) + mu (B n) :=
          by rw [hFinAdd n]
  have hPartialSums : Tendsto
      (fun n => sum k in Finset.range n, mu (A k))
      atTop (NHDS (mu (iUnion i, A i))) := by
    have h1 : forall n,
        (sum k in Finset.range n, mu (A k)) =
        mu (iUnion i, A i) - mu (B n) := by
      intro n
      have := hKey n
      have h2 : mu (iUnion i, A i) - mu (B n) =
          sum k in Finset.range n, mu (A k) +
          mu (B n) - mu (B n) := by rw [<- this]
      simp at h2; exact h2.symm
    simp_rw [h1]
    have : Tendsto
        (fun n => mu (iUnion i, A i) - mu (B n))
        atTop (NHDS (mu (iUnion i, A i) - 0)) :=
      Tendsto.sub tendsto_const_nhds hBtends
    simp at this; exact this
  have hHasSum : HasSum (fun n => mu (A n))
      (mu (iUnion i, A i)) := by
    rw [NNReal.hasSum_iff_tendsto_nat]; exact hPartialSums
  exact hHasSum.tsum_eq.symm
\end{lstlisting}}

\paragraph{Gemini 3 Pro (Incorrect).} The following proof attempt follows a similar decomposition strategy but fails due to issues with the convergence argument in the metric space section.

\begin{lstlisting}[
  frame=single,
  label={lst:gemini-dembo},
  caption={Gemini 3 Pro response (incorrect): fails in the convergence step.}
]
by
  let B := fun n => iUnion k, A (n + k)

  have hB_meas : forall n, MeasurableSet (B n) := by
    intro n; apply MeasurableSet.iUnion; intro k;
    exact hA (n + k)

  have hB_anti : Antitone B := by
    intro m n hmn x hx
    simp [B] at hx |-
    rcases hx with <k, hk>
    use (n - m) + k
    have : n + k = m + ((n - m) + k) := by
      rw [add_assoc, Nat.add_sub_of_le hmn]
    rw [this] at hk
    exact hk

  have hB_inter : (iInter n, B n) = empty := by
    rw [Set.eq_empty_iff_forall_notMem]
    intro x hx
    simp at hx
    rcases hx 0 with <k0, hk0>
    rcases hx (k0 + 1) with <k1, hk1>
    have neq : k0 /= k0 + 1 + k1 := by linarith
    exact (hDisj neq).ne_of_mem hk0 hk1 rfl

  have h_lim : Tendsto (fun n => mu (B n)) atTop (NHDS 0) :=
    hCont B hB_meas hB_anti hB_inter

  let S := fun n => iUnion k in Finset.range n, A k
  have hS_meas : forall n, MeasurableSet (S n) :=
    fun n => MeasurableSet.biUnion
      (Finset.range n).countable_toSet (fun k _ => hA k)

  have h_decomp : forall n, (iUnion k, A k) = S n \cup B n := by
    intro n; ext x; simp [S, B]
    constructor
    * rintro <k, hk>
      by_cases h : k < n
      * left; use k, h
      * right; use k - n;
        rw [Nat.add_sub_of_le (le_of_not_lt h)]; exact hk
    * rintro (<k, hk_lt, hk> | <k, hk>)
      * use k, hk
      * use n + k, hk

  have h_disj_SB : forall n, Disjoint (S n) (B n) := by
    intro n; rw [Set.disjoint_left]
    intro x hx_S hx_B
    simp [S, B] at hx_S hx_B
    rcases hx_S with <k1, hk1_lt, hk1>
    rcases hx_B with <k2, hk2>
    have : k1 /= n + k2 := by linarith
    exact (hDisj this).ne_of_mem hk1 hk2 rfl

  have h_add_n : forall n,
      mu (iUnion k, A k) = mu (S n) + mu (B n) := by
    intro n; rw [h_decomp n]
    apply hAdd (S n) (B n) (hS_meas n) (hB_meas n)
      (h_disj_SB n)

  have h_sum_S : forall n,
      mu (S n) = sum k in Finset.range n, mu (A k) := by
    intro n
    induction n with
    | zero =>
      simp [S]
      have : mu empty = 0 := by
        have := hAdd empty empty MeasurableSet.empty
          MeasurableSet.empty (disjoint_empty empty)
        simp at this; exact this
      exact this
    | succ n ih =>
      rw [Finset.sum_range_succ, <- ih]
      have hS_disj : Disjoint (S n) (A n) := by
        rw [Set.disjoint_left]
        intro x hx_S hx_A
        simp [S] at hx_S
        rcases hx_S with <k, hk_lt, hk>
        exact (hDisj (ne_of_lt hk_lt)).ne_of_mem
          hk hx_A rfl
      have hS_union : S (n + 1) = S n \cup A n := by
        ext x; simp [S]; constructor
        * rintro <k, hk_lt, hk>
          rw [Nat.lt_succ_iff_lt_or_eq] at hk_lt
          cases hk_lt
          * left; use k; assumption; assumption
          * right; subst k; assumption
        * rintro (<k, hk_lt, hk> | hxn)
          * use k; constructor;
            apply Nat.lt_succ_of_lt hk_lt; assumption
          * use n; constructor;
            apply Nat.lt_succ_self; assumption
      rw [hS_union]
      rw [hAdd (S n) (A n) (hS_meas n) (hA n) hS_disj]

  have h_conv : Tendsto (fun n => mu (S n))
      atTop (NHDS (mu (iUnion k, A k))) := by
    apply Metric.tendsto_nhds.mpr
    intro eps heps
    rw [Metric.tendsto_nhds] at h_lim
    specialize h_lim eps heps
    rw [Filter.mem_atTop_sets] at h_lim
    rcases h_lim with <N, hN>
    use N; intro n hn; specialize hN n hn
    rw [dist_eq, Real.dist_eq, abs_eq_self.mpr]
    * rw [h_add_n n]; rw [dist_comm]
      rw [dist_eq_norm]; simp
      rw [<- NNReal.coe_add]
      rw [add_sub_cancel_left]; simp
      rw [dist_eq_norm] at hN; simp at hN; exact hN
    * simp

  symm; apply HasSum.tsum_eq
  rw [NNReal.hasSum_iff_tendsto_nat]
  simp_rw [<- h_sum_S]; exact h_conv
\end{lstlisting}

Both proofs follow the same strategy, decompose $\mu(\bigcup_i A_i)$
into a finite partial sum and a vanishing tail, but differ in how they
handle the convergence step. Claude Opus 4.5 (Thinking) stays within
the \texttt{NNReal} API throughout, while Gemini 3 Pro attempts to use
real-valued metric lemmas (\texttt{Metric.tendsto\_nhds},
\texttt{dist\_eq}) on \texttt{NNReal} expressions, causing type
mismatches that the Lean elaborator cannot resolve.

\subsection{Tool guidance prompt}\label{app:prompts}

\begin{figure}[H]
\centering
\begin{minipage}{0.85\linewidth}
\scriptsize
\begin{framed}
\begin{verbatim}
TOOL_GUIDANCE_TEMPLATE = """BUDGET: {max_turns} turns total.
This is the total number of turns you have to submit your
final Lean 4 proof for the theorem. Each turn allows UP TO
6 TOOL CALLS MAX (calls beyond 6 are dropped without warning).

CRITICAL SCORING RULE: You get 0 points unless you call
submit_proof. No submission implies automatic failure.

Tools:
- lean_run_code: Test proofs and iterate on errors
- lean_loogle: Search for existing lemmas and definitions
  from Mathlib (use this tool carefully and efficiently)
- submit_proof: Submit final proof - REQUIRED TO GET ANY POINTS

IMPORTANT - Tool call strategy:
- You can make up to 6 tool calls per turn. Use them efficiently.
- Example: In one turn, you could call lean_loogle twice AND
  lean_run_code 4 times (6 total).
- Or call lean_run_code 6 times to test multiple proof
  variations at once.
- On your final turn, include submit_proof among your tool calls!

Be smart about how you use the tools; here are some tips:
- Submit your BEST attempt even if imperfect - an imperfect
  submission beats no submission
- If you have less than a few turns left, try to STOP exploring
  and call submit_proof as soon as possible
- If you are confident about your proof early on, call
  submit_proof immediately
- Never exhaust all {max_turns} turns without calling
  submit_proof
"""
\end{verbatim}
\end{framed}
\end{minipage}
\caption{Tool guidance template appended to the user prompt.}
\label{fig:tool-guidance}
\end{figure}

\end{document}